\documentclass[sn-vancouver,Numbered]{sn-jnl}

\usepackage{graphicx}%
\usepackage{threeparttable}
\usepackage{multirow}%
\usepackage{amsmath,amssymb,amsfonts}%
\usepackage{amsthm}%
\usepackage{mathrsfs}%
\usepackage[title]{appendix}%
\usepackage{xcolor}%
\usepackage{textcomp}%
\usepackage{manyfoot}%
\usepackage{booktabs}%
\usepackage{algorithm}%
\usepackage{algorithmicx}%
\usepackage{algpseudocode}%
\usepackage{listings}%
\usepackage{bm}
\usepackage[caption=false,labelformat=simple,font=footnotesize]{subfig}

\begin{document}

\title[Article Title]{Survey on Remote Sensing Scene Classification: From Traditional Methods to Large Generative AI Models}

\author*[1]{\fnm{Qionghao} \sur{Huang}}\email{qhhuang@m.scnu.edu.cn}
\author[1]{\fnm{Can} \sur{Hu}}\email{Sarahu@zjnu.edu.cn}

\affil*[1]{\orgname{Zhejiang Normal University}, \orgdiv{Zhejiang Key Laboratory of Intelligent Education Technology and Application}, \orgaddress{\city{Jinhua}, \postcode{321004}, \country{China}}}

\abstract{Remote sensing scene classification has experienced a paradigmatic transformation from traditional handcrafted feature methods to sophisticated artificial intelligence systems that now form the backbone of modern Earth observation applications. This comprehensive survey examines the complete methodological evolution, systematically tracing development from classical texture descriptors and machine learning classifiers through the deep learning revolution to current state-of-the-art foundation models and generative AI approaches. We chronicle the pivotal shift from manual feature engineering to automated hierarchical representation learning via convolutional neural networks, followed by advanced architectures including Vision Transformers, graph neural networks, and hybrid frameworks. The survey provides in-depth coverage of breakthrough developments in self-supervised foundation models and vision-language systems, highlighting exceptional performance in zero-shot and few-shot learning scenarios. Special emphasis is placed on generative AI innovations that tackle persistent challenges through synthetic data generation and advanced feature learning strategies. We analyze contemporary obstacles including annotation costs, multimodal data fusion complexities, interpretability demands, and ethical considerations, alongside current trends in edge computing deployment, federated learning frameworks, and sustainable AI practices. Based on comprehensive analysis of recent advances and gaps, we identify key future research priorities: advancing hyperspectral and multi-temporal analysis capabilities, developing robust cross-domain generalization methods, and establishing standardized evaluation protocols to accelerate scientific progress in remote sensing scene classification systems.}


\maketitle

\section{Introduction}
\label{sec:intro}
Remote sensing (RS) scene classification is a pivotal task in Earth observation, enabling the categorization of satellite and aerial imagery into meaningful semantic classes for enhanced environmental monitoring, urban planning, and disaster management~\cite{cheng2017remote,thapa2023deep}. This survey explores the progression from traditional handcrafted feature methods to advanced deep learning, foundation models, and generative AI techniques, highlighting key methodologies, persistent challenges, and emerging future directions~\cite{cheng2020remote,dai2024multi}.

\subsection{Background and Motivation}

Remote sensing technologies have evolved dramatically from early aerial photography to advanced multispectral, hyperspectral, SAR, and LiDAR systems~\cite{liu2024a}. This progression, driven by sensor advancements and satellite constellations, has enabled higher spatial, spectral, and temporal resolutions~\cite{elachi2021introduction}. The integration of AI with RS has facilitated real-time data analysis and enhanced Earth observation capabilities~\cite{moreno2024federated}. Global RS data analysis capabilities have expanded substantially, reflecting increasing availability of diverse sources including high-resolution optical satellites, UAVs, and ground-based sensors~\cite{dritsas2025remote}. Future trends emphasize automated digital processes where RS data becomes integral to decision-making, supported by explosive growth in EO capabilities and small satellite constellations~\cite{li2023autonomous,siddique2024small}. Legal frameworks are evolving to address data processing with AI advancements, ensuring ethical utilization~\cite{paolanti2024ethical}.

Scene classification involves categorizing entire images into predefined land-use or land-cover classes, enabling actionable insights from vast EO datasets~\cite{tseng2023ga}. Its importance spans urban planning for monitoring land-use changes and infrastructure development; disaster management through rapid damage assessment and hazard mapping; and environmental monitoring, including ecosystem health evaluation and climate change tracking~\cite{zhang2022artificial}. In disaster scenarios, RS technologies using thermal, infrared, and microwave sensors provide timely data for response strategies~\cite{janga2023review}. Urban applications support sustainable development by detecting environmental shifts and integrating with spatial big data~\cite{yin2021integrating}. Overall, RS scene classification bridges raw data to practical outcomes in agriculture, natural resource management, and policy formulation~\cite{liu2020local}.

\subsection{Scope and Contributions of the Survey}

This survey comprehensively covers RS scene classification methodologies from traditional handcrafted feature extraction and classical machine learning classifiers through deep learning paradigms like CNNs and vision transformers, to cutting-edge large generative AI models and remote sensing foundation models (RSFMs)~\cite{lu2024ai}. It encompasses GANs, VAEs, diffusion models, and multimodal vision-language models tailored for RS tasks~\cite{dash2023review}. Special attention is given to recent foundation models released after 2021, including self-supervised learning strategies and contrastive approaches for handling large-scale RS data with minimal annotations~\cite{sun2022ringmo}.

While prior surveys have made important contributions to the field, this work addresses several gaps they leave open. Cheng et al.~\cite{cheng2020remote} provided a systematic review of deep learning methods up to 2020 but predates the emergence of vision-language models, large-scale RSFMs, and generative AI approaches. Thapa et al.~\cite{thapa2023deep} conducted a meta-analysis of CNN-, ViT-, and GAN-based architectures published through 2023, yet their scope does not extend to billion-parameter foundation models such as SkySense~\cite{guo2024skysense}, self-supervised pretraining strategies such as RingMo~\cite{sun2022ringmo} and Scale-MAE~\cite{reed2023scale}, or federated and edge-computing deployment paradigms. The present survey uniquely integrates all of these developments into a unified methodological narrative, covering the complete evolution from handcrafted descriptors to state-of-the-art generative and foundation model approaches, while explicitly analyzing emerging challenges in interpretability, adversarial robustness, sustainable AI, and standardized benchmarking that are absent or only partially treated in existing reviews.

Key insights include superior performance of transformer-based and foundation models in addressing challenges like intra-class variability and domain shifts, with comparative analyses highlighting accuracy improvements in zero-shot and few-shot classification. Identified gaps involve data imbalance, annotation costs, and multimodal integration needs, particularly in hyperspectral and multi-temporal processing~\cite{kaul2023literature}. The survey evaluates recent advances, including RSFMs and generative AI integrations, revealing trends toward hybrid architectures and brain-inspired models for enhanced geospatial understanding~\cite{jiao2023brain}. Contributions include synthesizing emerging trends like edge computing and federated learning while identifying benchmarking and scalability opportunities~\cite{tombe2023remote}.

\subsection{Organization of the Survey}

\begin{figure}[!t]
    \centering
    \includegraphics[width=0.99\textwidth]{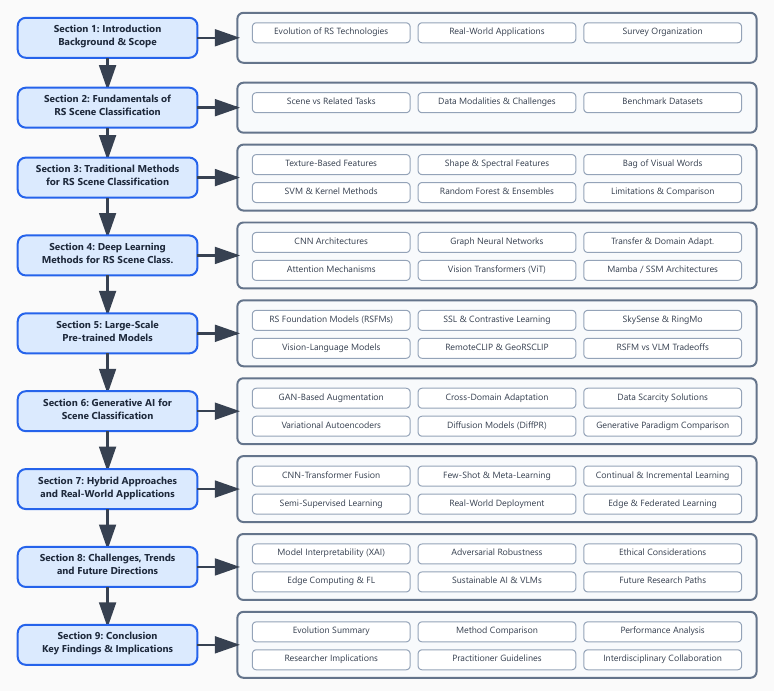}
    \caption{Structure of the Survey}
    \label{fig:strct}
\end{figure}

The survey is structured (shown in Figure~\ref{fig:strct}) as follows: Section 2 establishes fundamentals, including definitions, challenges, benchmark datasets, and evaluation metrics. Section 3 reviews traditional methods covering handcrafted features and classical classifiers. \textcolor{black}{Section 4 provides a comprehensive treatment of deep learning methods, covering CNNs, graph neural networks, transfer learning and domain adaptation, attention mechanisms, Vision Transformers, and state space models (Mamba). Section 5 focuses on large-scale pre-trained models, including Remote Sensing Foundation Models (RSFMs) and Vision-Language Models (VLMs).} Section 6 investigates generative AI applications covering GANs, VAEs, and diffusion models. Section 7 presents advanced hybrid approaches and real-world applications, including ensemble methods and federated learning. Section 8 addresses current challenges and future directions, emphasizing interpretability, ethics, and emerging trends. Section 9 concludes with key findings and implications for advancing the field. \textcolor{black}{Figure~\ref{fig:taxonomy} provides an overview of the methodological taxonomy and developmental timeline, illustrating how each paradigm evolved in response to the limitations of its predecessors.}

\textcolor{black}{\begin{figure}[!t]
    \centering
    \includegraphics[width=0.99\textwidth]{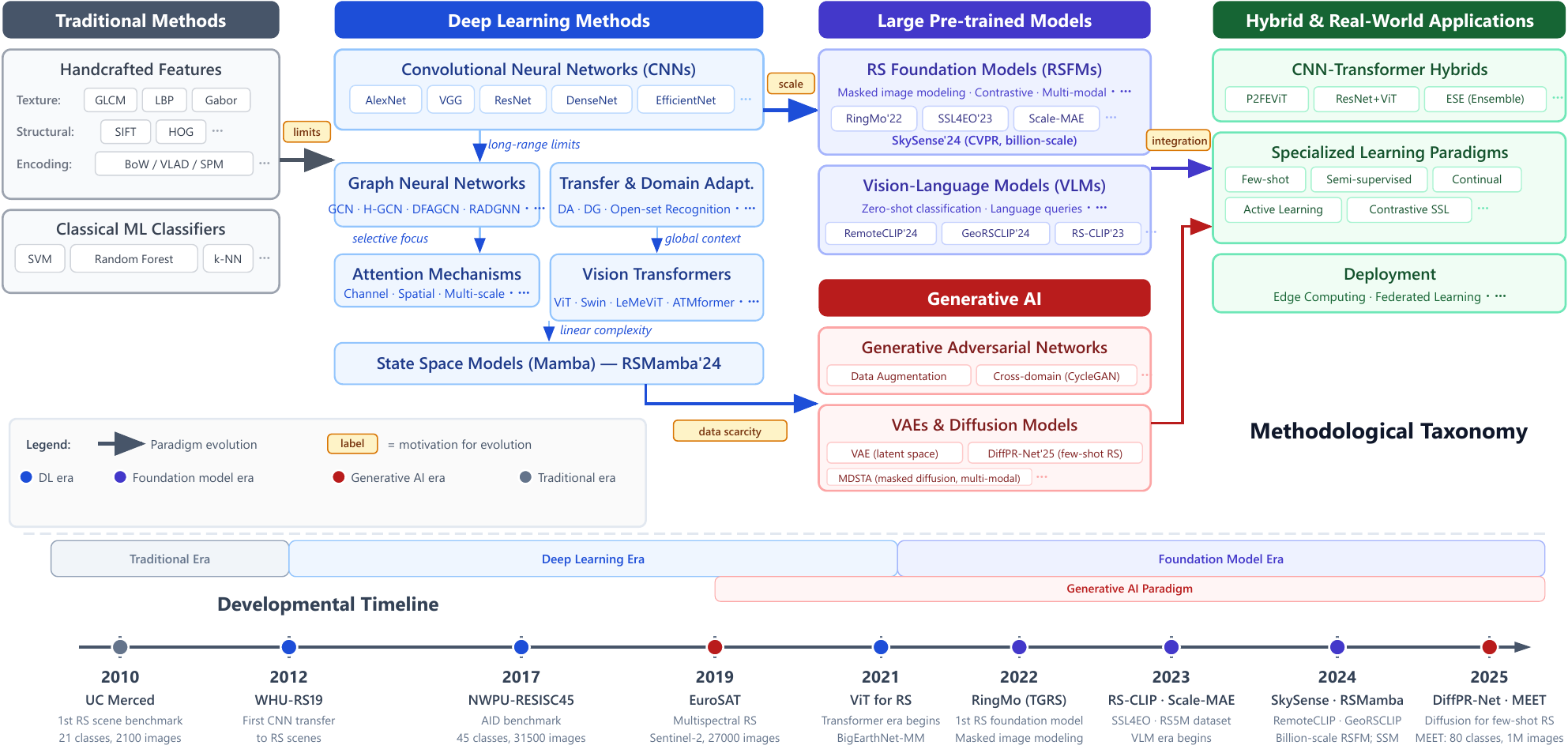}
    \caption{\textcolor{black}{Methodological taxonomy and developmental timeline of remote sensing scene classification. The diagram traces the evolution from traditional handcrafted feature methods (e.g., BoW, VLAD, spectral/texture descriptors) through deep learning architectures (CNNs, GNNs, Attention Mechanisms, Vision Transformers, Mamba) to large-scale pre-trained models (RSFMs, VLMs) and generative AI approaches (GANs, VAEs, Diffusion Models), as well as hybrid and real-world application strategies. Branch connections indicate how later paradigms were motivated by the limitations of earlier approaches, while timeline annotations mark the approximate periods of prominence for each category.}}
    \label{fig:taxonomy}
\end{figure}}

\section{Fundamentals of Remote Sensing Scene Classification}
\label{sec:frssc}
Remote sensing scene classification is a foundational task in Earth observation that involves assigning semantic labels to entire remote sensing images or scenes based on their holistic content rather than individual pixels or objects~\cite{tombe2023remote}. It serves as a critical component for automated interpretation of large-scale geospatial data, enabling applications in land-use monitoring, environmental assessment, urban planning, and disaster management~\cite{cheng2020remote}.

\subsection{Definition and Key Concepts}

Remote sensing scene classification aims to categorize entire images based on their holistic semantic content and spatial context, distinguishing it from pixel-level analysis or object-specific detection tasks~\cite{tombe2023remote}. 
The fundamental challenge lies in extracting semantic features from images with significant intra-class variation, such as agriculture, airports, commercial areas, and residential areas~\cite{stojnic2021self}.

\textbf{Scene Classification vs. Related Remote Sensing Tasks.}
Scene classification occupies a distinct position within the remote sensing image analysis pipeline, focusing on holistic understanding rather than localized analysis~\cite{kaul2023literature}. 
Different from recognizing the commercial scene or residential scene in an image, semantic segmentation networks can delineate the boundaries of each land object in the scene and achieve densely pixel-wise predictions both semantically and locationally~\cite{pires2019convolutional}. 
Object detection in remote sensing identifies and localizes specific objects within scenes using bounding boxes, requiring precise spatial localization capabilities~\cite{thapa2023deep}. While scene classification provides high-level categorical understanding suitable for broad-scale monitoring and environmental assessment, object detection and segmentation tasks offer detailed spatial information but typically require substantially more annotated training data and greater computational resources~\cite{chen2023unified}.

\textbf{Types of Remote Sensing Data and Modalities.}
Remote sensing data encompasses diverse modalities, each providing unique spectral, spatial, and temporal characteristics essential for comprehensive scene understanding~\cite{agrawal2019comparative}. 
Optical imagery, including visible RGB and multispectral data from satellites such as Landsat and Sentinel-2, captures reflected solar radiation across visible and near-infrared spectral bands, enabling detailed analysis of vegetation health, water quality, and surface materials~\cite{elachi2021introduction}. 
Synthetic Aperture Radar (SAR) systems provide all-weather, day-night imaging capabilities through active microwave sensing, offering unique advantages for penetrating clouds and vegetation while giving information about surface roughness and structural properties~\cite{sabins2020remote}. 
Hyperspectral sensors acquire data across hundreds of narrow, contiguous spectral bands, enabling detailed material discrimination and chemical composition analysis~\cite{chen2020using}. 
Light Detection and Ranging (LiDAR) systems generate precise 3D point clouds through laser pulse measurements, providing accurate elevation models and structural information~\cite{kahraman2021comprehensive}.
The integration and fusion of multiple sensor modalities has become increasingly crucial for robust scene classification, as different sensors provide complementary information that can significantly improve classification accuracy across diverse environmental conditions.
Table~\ref{tab:rs_modalities} summarizes the key characteristics and applications of these remote sensing data modalities.

\begin{table}[!ht]
\centering
\footnotesize
\caption{Remote Sensing Data Modalities for Scene Classification}
\label{tab:rs_modalities}
\begin{tabular}{|l|l|l|}
\hline
\textbf{Data Modality} & \textbf{Key Characteristics} & \textbf{Main Applications} \\
\hline
Optical Imagery~\cite{elachi2021introduction} & Visible/near-infrared bands & Vegetation, water, surface materials \\
\hline
SAR~\cite{sabins2020remote} & All-weather microwave sensing & Surface roughness, structural properties \\
\hline
Hyperspectral~\cite{chen2020using} & Hundreds of spectral bands & Material discrimination, composition \\
\hline
LiDAR~\cite{kahraman2021comprehensive} & 3D point clouds & Elevation models, structural information \\
\hline
\end{tabular}
\end{table}

\subsection{Fundamental Technical Challenges in Remote Sensing Scene Classification}

Remote sensing scene classification faces intrinsic technical challenges due to the complex nature of geospatial imagery and the unique characteristics of Earth observation data~\cite{cheng2017remote}.

\textbf{Intra-Class Variability and Inter-Class Similarity.}
High intra-class variance occurs when the same class scene images are commonly taken at varying angles, scales, and viewpoints, requiring well-designed computer vision approaches that can extract the same pattern features regardless of their variations~\cite{zhao2021mgml}. For example, urban scenes may range from dense metropolitan areas to suburban neighborhoods, while agricultural scenes encompass diverse crop types and farming practices across different regions~\cite{dai2024multi}.
Low inter-class variance presents challenges when scene images appear similar, such as agriculture and forest, or dense residential and suburban areas~\cite{dai2024multi}.

\textbf{Complex Spatial Organization and Multi-Scale Issues.}
Remote sensing images present formidable classification challenges due to their complex spatial organization, high inter-class similarity, and significant intra-class variability~\cite{wang2022self}. Unlike typical natural images where objects dominate the image space, RS scenes frequently contain multiple objects dispersed across cluttered backgrounds, while the imaging angle introduces substantial background noise that obscures key features~\cite{luo2025multisensor}. This requires classification models to capture both fine-grained local features and broad contextual information effectively.

\textbf{Environmental and Sensor-Related Factors.}
Environmental conditions significantly impact image quality and appearance, creating additional complexity for robust classification systems~\cite{luo2025multisensor}. Atmospheric effects, including varying illumination conditions, seasonal changes affecting vegetation phenology, and weather conditions, can substantially alter spectral signatures and visual appearance of land cover types~\cite{thiruchittampalam2025evaluation}.
Sensor-related challenges include radiometric variations between different platforms, geometric distortions, and sensor noise characteristics that can affect classification performance when models are applied across different data sources~\cite{misra2022remote}.

\subsection{Benchmark Datasets and Evaluation Frameworks}

Standardized benchmark datasets and evaluation metrics are essential for objective comparison of scene classification methods and tracking progress in the field~\cite{sumbul2019bigearthnet}. The evolution of benchmark datasets reflects the progression from small-scale academic datasets to large-scale operational challenges.

\textbf{Classical and Foundational Datasets.}
Early benchmark datasets established fundamental evaluation frameworks for algorithm development and comparison~\cite{cheng2017remote}. The UCMerced dataset, published in 2010, is one of the earliest datasets with only 2,100 images distributed over 21 classes, making it unsuitable for deep learning models, which need large amounts of data~\cite{yang2010bag}. The Aerial Image Dataset (AID) expanded the scale with 10,000 aerial images across 30 scene categories, emphasizing diverse geographic locations and imaging conditions~\cite{xia2017aid}. NWPU-RESISC45 further increased complexity with 31,500 images spanning 45 scene classes collected from multiple countries and regions, establishing a more challenging benchmark for evaluating generalization capabilities~\cite{cheng2017remote}. EuroSAT leverages Sentinel-2 multispectral satellite imagery with 27,000 labeled samples across 10 land cover classes, representing the transition toward operational satellite-based classification scenarios~\cite{helber2019eurosat}. BigEarthNet-MM represents a significant advance as a large-scale multimodal dataset combining Sentinel-1 SAR and Sentinel-2 optical imagery, supporting multi-label scene classification research with over 590,000 image patches~\cite{sumbul2021bigearthnet}.

\begin{table}[!ht]
\centering
\footnotesize
\caption{\textcolor{black}{Comprehensive Summary of Remote Sensing Scene Classification Benchmark Datasets by Data Modality}}
\label{tab:rs_benchmark_datasets}
\textcolor{black}{\begin{tabular}{|l|l|l|l|l|l|l|}
\hline
\textbf{Dataset} & \textbf{Images} & \textbf{Classes} & \textbf{Resolution} & \textbf{Modality} & \textbf{Balance} & \textbf{Year} \\
\hline
\multicolumn{7}{|c|}{\textbf{RGB Optical Datasets}} \\
\hline
UC Merced~\cite{yang2010bag} & 2,100 & 21 & 256$\times$256 px & RGB Aerial & Balanced & 2010 \\
\hline
WHU-RS19 & 1,005 & 19 & 600$\times$600 px & RGB Aerial & Imbalanced & 2012 \\
\hline
AID~\cite{xia2017aid} & 10,000 & 30 & 600$\times$600 px & RGB Aerial & Imbalanced & 2017 \\
\hline
PatternNet & 30,400 & 38 & 256$\times$256 px & RGB Satellite & Balanced & 2018 \\
\hline
NWPU-RESISC45~\cite{cheng2017remote} & 31,500 & 45 & 256$\times$256 px & RGB Aerial & Balanced & 2017 \\
\hline
MEET~\cite{li2025meet} & 1,030,000 & 80 & 256$\times$256 px & RGB at 1.0m & Imbalanced & 2025 \\
\hline
\multicolumn{7}{|c|}{\textbf{Multispectral Datasets}} \\
\hline
EuroSAT~\cite{helber2019eurosat} & 27,000 & 10 & 64$\times$64 px & Sentinel-2 MS & Imbalanced & 2019 \\
\hline
\multicolumn{7}{|c|}{\textbf{Hyperspectral Datasets}} \\
\hline
Indian Pines & 21,025 & 16 & 145$\times$145 px & AVIRIS HSI & Imbalanced & 1992 \\
\hline
Pavia University & 207,400 & 9 & 610$\times$340 px & ROSIS HSI & Imbalanced & 2001 \\
\hline
Houston 2013 & 15,029 & 15 & 349$\times$1905 px & ITRES HSI & Imbalanced & 2013 \\
\hline
\multicolumn{7}{|c|}{\textbf{Multi-modal and Large-Scale Datasets}} \\
\hline
BigEarthNet-MM~\cite{sumbul2021bigearthnet} & 590,326 & 19 (multi-label) & 120$\times$120 px & S1 SAR + S2 MS & Imbalanced & 2021 \\
\hline
RS5M~\cite{zhang2024rs5m} & 5,000,000 & Text-paired & Variable & Multi-source & N/A & 2023 \\
\hline
\end{tabular}}
\end{table}

\textcolor{black}{The expanded Table~\ref{tab:rs_benchmark_datasets} systematically categorizes benchmark datasets by data modality, including RGB optical, multispectral, hyperspectral, and multi-modal datasets. Notably, class imbalance is prevalent across most datasets: in Indian Pines, the largest class (Corn-notill, 1,428 samples) contains over 25 times more samples than the smallest class (Oats, 20 samples); in EuroSAT, Industrial Buildings and Residential areas are significantly underrepresented compared to Crop categories. This persistent class imbalance poses ongoing challenges for fair model evaluation and has motivated research into balanced sampling strategies, cost-sensitive learning, and synthetic data augmentation.}

\textbf{Large-Scale and Contemporary Datasets.}
Recent dataset developments address the scalability and diversity requirements of modern foundation models and operational applications. The RS5M dataset comprises over 5 million remote sensing images explicitly designed for pretraining large-scale foundation models, representing a paradigm shift toward self-supervised learning approaches~\cite{zhang2024rs5m}. The MEET dataset includes over 1.03 million fine-grained geospatial images with comprehensive zoom-level variations, supporting research in scale-invariant classification and multi-resolution analysis~\cite{li2025meet}. Table~\ref{tab:rs_benchmark_datasets} demonstrates the clear evolution from classical datasets with thousands of images to contemporary datasets with millions of samples, reflecting the increasing data requirements and computational capabilities of modern deep learning approaches.

\textbf{Evaluation Metrics and Performance Assessment.}
Standardized evaluation metrics ensure objective and comparable assessment of classification performance across different methods and datasets. Overall Accuracy measures the proportion of correctly classified samples relative to total test samples, while F1-score balances precision and recall, particularly for datasets with class imbalance, and the Kappa coefficient accounts for chance agreement to provide more robust measures than simple accuracy~\cite{congalton1991review}. Per-class accuracy metrics offer detailed insights into model performance across different scene types, while confusion matrix visualization enables comprehensive error analysis~\cite{lewis2001generalized}. Modern evaluation frameworks increasingly incorporate specialized scenarios such as few-shot learning settings and semi-supervised learning benchmarks that evaluate performance with partially labeled datasets~\cite{thapa2023deep}.

\section{Traditional Methods for Remote Sensing Scene Classification}

Before the advent of deep learning, RS scene classification relied on handcrafted feature extraction combined with classical machine learning algorithms~\cite{cheng2017remote,thapa2023deep}. These manually designed descriptors captured texture, spectral, and structural characteristics but required extensive domain expertise~\cite{tombe2023remote}. While effective for simple scenarios and smaller datasets, traditional approaches faced significant limitations with the complex intra-class variability and inter-class similarity inherent in RS imagery.

\subsection{Handcrafted Feature Extraction Techniques}
Traditional methods employed mathematical descriptors operating at local and global image levels to characterize scene content for classification.

\textbf{Texture-Based Feature Descriptors.} 
Gray-Level Co-occurrence Matrix (GLCM) analyzes spatial relationships between pixel pairs by computing the frequency of co-occurring intensity values at specified distances and orientations, extracting statistical measures including contrast, correlation, energy, and homogeneity to characterize texture patterns~\cite{haralick2007textural,mirzapour2015fast}. Local Binary Patterns (LBP) encode local texture information by comparing each pixel's intensity with its neighborhood, generating binary codes that provide rotation-invariant texture descriptions robust to illumination variations~\cite{ojala2002multiresolution,anwer2018binary}. Gabor filters apply multi-scale and multi-orientation filtering using Gaussian-modulated sinusoidal functions, effectively capturing texture frequency and directional information across different scales~\cite{daugman1985uncertainty}.

\textbf{Structural and Spectral Feature Descriptors.} 
Scale-Invariant Feature Transform (SIFT) detects distinctive keypoints with descriptors that remain invariant to scale, rotation, and illumination changes, proving particularly valuable for RS scene classification when combined with spatial pyramid matching frameworks~\cite{lowe2004distinctive}. Histogram of Oriented Gradients (HOG) captures structural information through gradient orientation distributions, effectively describing geometric patterns such as buildings, roads, and linear infrastructure in aerial imagery~\cite{dalal2005histograms,mehmood2022remote}. Spectral feature descriptors, including color histograms and band ratio indices, represent global intensity distributions and spectral characteristics, particularly valuable for distinguishing vegetation, water bodies, and soil types~\cite{misra2022remote}.

\textbf{Bag-of-Visual-Words Framework.} 
The Bag-of-Visual-Words (BoVW) model adapts document analysis techniques to image classification by treating local features as visual words and images as documents, creating discriminative vocabularies through unsupervised clustering methods such as k-means~\cite{csurka2004visual,zhu2016bag}. Enhanced BoVW variants incorporate spatial information through spatial pyramid matching, which preserves spatial relationships by partitioning images into multi-resolution grids~\cite{lazebnik2006beyond}. Advanced approaches address rotation invariance and background interference through saliency-guided feature selection and region-based covariance descriptors~\cite{chen2022bag}.

\subsection{Classical Machine Learning Classifiers}
Traditional classification algorithms operated on extracted handcrafted features to perform scene-level categorization, with several approaches proving particularly effective for RS applications.

\textbf{Support Vector Machines and Ensemble Methods.} 
Support Vector Machines (SVMs) construct optimal separating hyperplanes in high-dimensional feature spaces, demonstrating particular effectiveness for RS scene classification due to their ability to handle high-dimensional feature vectors and maintain performance with limited training samples~\cite{burges1998tutorial,mountrakis2011support}. Random Forest classifiers aggregate predictions from multiple decision trees trained on bootstrapped samples, effectively reducing overfitting while handling the noisy and heterogeneous features characteristic of RS imagery~\cite{breiman2001random,belgiu2016random}.

\textbf{Additional Classical Approaches.} 
Maximum Likelihood Classifiers assume Gaussian distributions for feature vectors within each scene class, providing probabilistic classification decisions based on statistical modeling~\cite{richards2006remote}. k-Nearest Neighbors (k-NN) performs classification through majority voting among the k most similar training samples, offering non-parametric classification suitable for complex feature distributions~\cite{cover1967nearest}. Ensemble methods, including AdaBoost and gradient boosting, combine multiple weak classifiers to achieve improved performance, with studies demonstrating classification accuracies exceeding 90\% on standard RS scene classification benchmarks~\cite{dou2018remote}.

\subsection{Limitations of Traditional Approaches}
Traditional methods face critical scalability and generalization issues, with handcrafted features like GLCM requiring intensive computation that scales poorly with modern high-resolution datasets~\cite{sheykhmousa2020support}. They struggle with cross-domain adaptation across different geographic regions, sensor types, or acquisition conditions~\cite{tuia2009active,demir2015hashing}. Additionally, traditional methods require extensive domain expertise and manual feature engineering, introducing subjective biases affecting reproducibility~\cite{mellor2015exploring,foody2002status}, while struggling to capture complex spatial relationships and multi-scale patterns essential for accurate classification in complex scenes~\cite{cheng2017remote}.

\textcolor{black}{In summary, traditional methods offer notable advantages in interpretability and low computational cost, making them suitable for resource-constrained settings with small, simple datasets. However, they are fundamentally limited by their reliance on manually designed features that cannot capture the hierarchical, multi-scale representations needed for complex RS scenes. The transition to deep learning, discussed in the next section, addresses these limitations through automatic feature learning at the cost of increased data requirements and reduced model transparency.}

\section{\textcolor{black}{Deep Learning Methods for Remote Sensing Scene Classification}}
\textcolor{black}{Deep learning has revolutionized RS scene classification by enabling automatic hierarchical feature extraction through end-to-end learning, significantly outperforming traditional handcrafted feature methods~\cite{cheng2017remote,thapa2023deep}.}
\textcolor{black}{This section reviews the major deep learning paradigms for RS scene classification, covering CNNs, graph neural networks, transfer learning and domain adaptation, attention mechanisms, Vision Transformers, and state space models (Mamba), providing a unified treatment of all deep learning architectures before discussing large-scale pre-trained models in Section~5.}

\subsection{Convolutional Neural Networks for Scene Classification}

CNNs have become the cornerstone of RS scene classification by learning hierarchical feature representations~\cite{shakya2021parametric}. Through weight sharing and translation equivariance, they effectively capture multi-scale spatial patterns and handle the complex spatial organization and intra-class variability characteristic of RS scenes~\cite{alhichri2021classification}.

\begin{figure}[!t]
    \centering
    \includegraphics[width=0.99\textwidth]{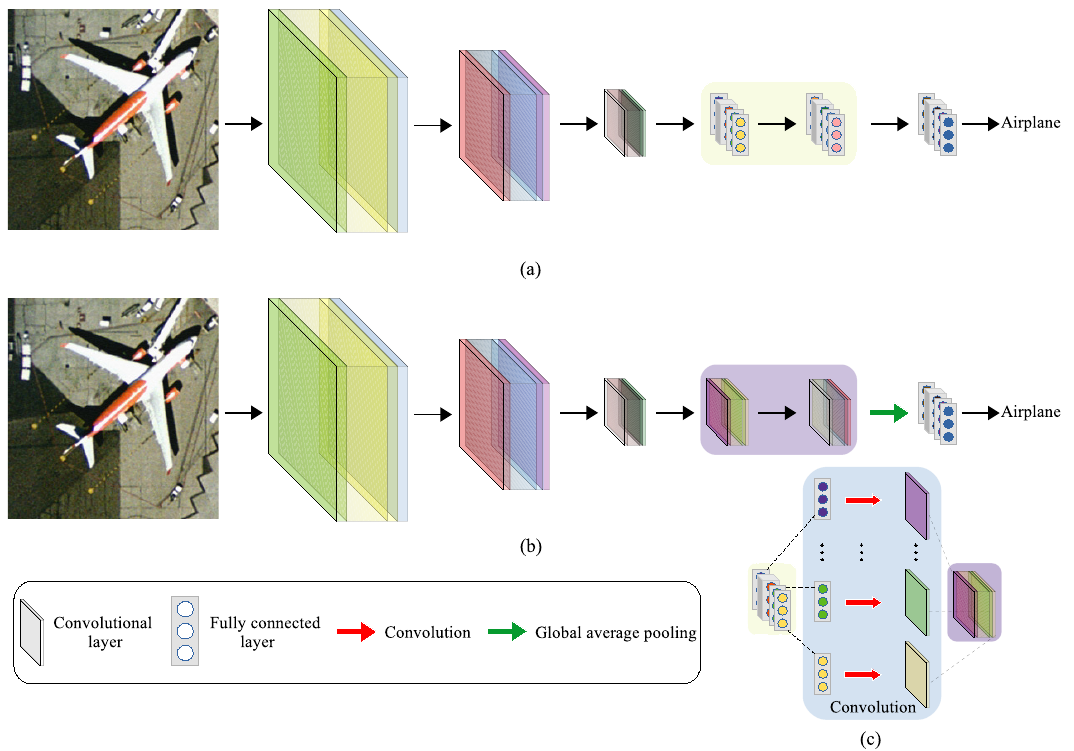} 
    \caption{Scale-Free CNN architecture: (a) traditional CNN with fixed input size, (b) SF-CNN enabling arbitrary input sizes through FCL convolution and global average pooling, (c) FCL convolution process~\cite{xie2019scale}.}
    \label{fig:sf_cnn_arch}
\end{figure}

\textbf{Early CNN Architectures for Remote Sensing.}
AlexNet adaptations were among the first successful deep learning applications in RS scene classification, with AlexNet-SPP-SS demonstrating superior performance on the UC Merced dataset~\cite{han2017pre}. 
VGG architectures showed particular promise due to their uniform structure and ability to capture fine-grained details, with VGG-based networks achieving competitive results on AID and NWPU-RESISC45 datasets~\cite{xu2020two}. 
Liu et al. proposed VGG-SA, replacing VGG-19's last four layers with self-attention blocks, showing improved performance on the AID dataset~\cite{liu2022scene}. 
Inception-based models brought multi-scale feature extraction capabilities to RS scene classification, with Scale-Free CNN addressing the limitation of fixed input sizes by converting fully connected layers to convolutional operations, enabling processing of arbitrary input dimensions (Figure~\ref{fig:sf_cnn_arch})~\cite{xie2019scale}. 
These architectures proved effective for RS scenes containing objects at varying scales and spatial arrangements.

\textbf{Advanced CNN Architectures for Scene Classification.}
ResNet architectures revolutionized deep RS scene classification by enabling much deeper networks through residual connections, consistently outperforming other architectures on standard benchmarks~\cite{alafandy2020using,adegun2023review}. 
The skip connections in ResNet effectively address the vanishing gradient problem while enabling the learning of identity mappings, crucial for preserving low-level spatial details in RS scenes.
DenseNet models introduced dense connectivity patterns that enhance feature reuse while reducing parameters, demonstrating superior performance on standard benchmarks with improved robustness to outliers~\cite{alafandy2022efficient,li2020classification}. 
EfficientNet brought compound scaling principles to RS scene classification, systematically balancing network depth, width, and resolution while maintaining computational efficiency~\cite{alosaimi2023self,alhichri2021classification}. 
STConvNeXt introduces a lightweight network designed explicitly for RS challenges, incorporating split-based mobile convolution modules with hierarchical structures and employing parameterized depthwise separable convolutions to reduce complexity while maintaining feature extraction capability~\cite{liu2025efficient}.

\subsection{Graph Neural Networks for Scene Classification}

\textcolor{black}{GNNs address CNN limitations by modeling long-range dependencies through graph structures, where nodes represent image regions and edges encode spatial or semantic relationships, enabling more effective handling of complex spatial organization in RS imagery~\cite{huang2025remote}.}

\textbf{Graph Convolutional Networks for Scene Understanding.}
Graph Convolutional Networks (GCNs) extend traditional convolution operations to graph-structured data, enabling the capture of both local and global spatial relationships inherent in remote sensing scenes~\cite{gao2021remote}. 
Gao et al. proposed H-GCN, which combines attention mechanisms with high-order graph convolution to focus on discriminative scene components while modeling class dependencies, demonstrating superior performance over traditional CNN approaches on UCM and NWPU-RESISC45 datasets~\cite{gao2021remote}. 
To address CNN limitations in capturing contextual relationships within scenes, Xu et al. developed DFAGCN, which treats image patches as graph nodes connected based on spatial proximity and feature similarity, achieving strong overall accuracy on NWPU-RESISC45~\cite{xu2022deep}.

\begin{figure}[!t]
    \centering
    \includegraphics[width=0.99\textwidth]{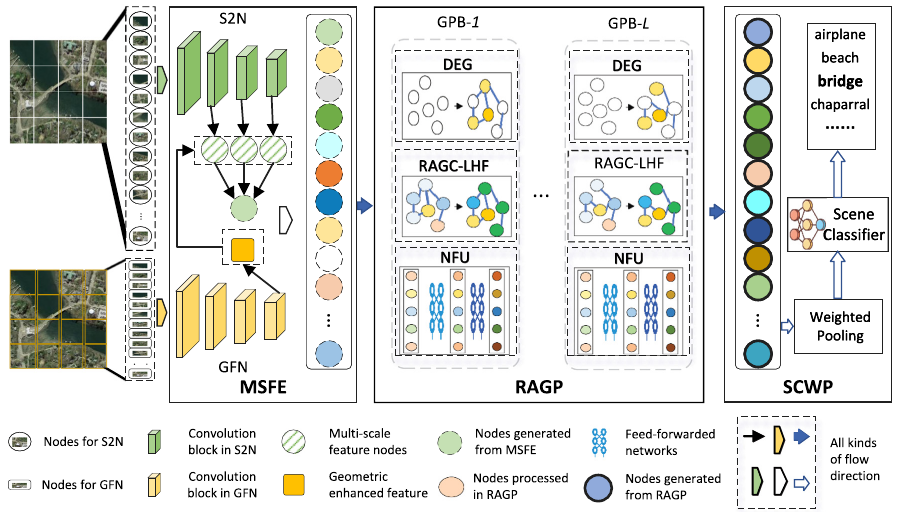} 
    \caption{The framework of RS-RADGNN~\cite{huang2025remote}.} 
    \label{fig:gcns}
\end{figure}

\textbf{Advanced Graph Network Approaches for Scene Classification.}
Huang et al. introduced RS-RADGNN with relation-aware dynamic graph construction that adaptively adjusts node connections based on learned scene representations~\cite{huang2025remote}, as shown in Figure~\ref{fig:gcns}. 
The approach incorporates multi-scale feature extraction and relation-aware dynamic graph convolution to capture important latent correlations between scene elements, addressing the dynamic nature of spatial relationships in diverse RS scenes.

Wang et al. proposed Comparison Graph Neural Networks (CGNN) to tackle large-scale variance within specific scene categories by constructing comparison graphs based on sample features extracted by CNN models~\cite{wang2022cgnn}.
This approach enables the model to learn from inter-sample relationships rather than relying solely on individual scene features.

\subsection{Transfer Learning and Domain Adaptation}
Transfer learning addresses the challenge of limited labeled RS data by leveraging knowledge from pre-trained models~\cite{rouba2023improving}. However, the domain gap between natural images and RS imagery requires careful adaptation across sensors, geographic regions, and temporal variations~\cite{zheng2022partial}.

\textbf{Pre-Trained Models and Fine-Tuning Strategies.}
Pre-trained models from ImageNet have proven surprisingly effective for RS scene classification despite domain differences, with the transferability stemming from shared low-level features like edges and textures~\cite{adegun2023review}.
Comprehensive studies comparing VGG16, ResNet50, DenseNet121, and EfficientNet-B3 have shown consistent improvements when using ImageNet pre-trained weights, with fine-tuning typically yielding substantial performance gains on standard RS benchmarks.
Layer-wise fine-tuning strategies have been developed specifically for RS scene classification applications. The most effective approach involves freezing early convolutional layers that capture low-level features while fine-tuning later layers and classification heads to adapt to RS-specific semantic patterns~\cite{zhang2020transfer}.
This selective fine-tuning approach significantly reduces training time while achieving comparable or superior accuracy to training from scratch~\cite{xu2023universal}.

\textcolor{black}{\textbf{Domain Adaptation.}}
\textcolor{black}{Domain adaptation (DA) methods address the distribution shift between source and target domains that commonly arises from differences in sensors, geographic regions, seasonal variations, and acquisition conditions in RS imagery~\cite{zheng2022partial}. Unsupervised DA methods align source and target feature distributions without requiring target labels. Adversarial-based approaches, such as domain adversarial neural networks, learn domain-invariant features through minimax training between a feature extractor and a domain discriminator, achieving strong cross-domain RS scene classification performance. Multi-source domain adaptation methods such as SSDAN (Semi-Supervised Domain Adaptation Network) using EfficientNet-B3 handle variations across multiple source domains simultaneously~\cite{lasloum2021ssdan}. Domain-invariant feature learning techniques focus on learning representations that are robust to domain variations while preserving discriminative information for scene classification~\cite{li2024semisupervised}. Ji et al. proposed DA-IDANet, which incorporates domain adaptive constraints to mitigate pseudo-changes caused by sensor differences and imaging conditions, demonstrating effective domain alignment through an interactive differential attention module~\cite{ji2024daidanet}. Open-set domain adaptation further extends this paradigm by jointly handling domain shift and unknown categories; Zheng et al. developed MAOSDAN, combining multi-adversarial learning with attention-aware open-set back-propagation to distinguish unknown from known samples while adapting between domains, achieving state-of-the-art results on cross-domain RS scene classification benchmarks~\cite{zheng2024openset}.}

\textcolor{black}{\textbf{Domain Generalization.}}
\textcolor{black}{Unlike domain adaptation, domain generalization (DG) aims to train models that generalize to completely unseen target domains without any access to target data during training---a more challenging yet practically important setting for RS applications where deployment environments cannot be anticipated~\cite{han2024mscdg}. Multi-source collaborative DG methods leverage the homogeneity and heterogeneity characteristics of multiple source domains through data-aware adversarial augmentation and model-aware multilevel diversification, effectively bridging cross-scene domain gaps~\cite{han2024mscdg}. Style and content separation approaches disentangle domain-specific style information from domain-invariant content features, enabling models to learn generalizable representations that transfer across different geographic regions and sensor types~\cite{zhu2023scsn}. For hyperspectral image classification, spectral shifts between different scenes pose particularly severe challenges. Chen et al. proposed DAMS, which leverages natural language to guide visual models for single-domain generalization by disentangling modality-shared semantics from modality-specific attributes~\cite{chen2025dams}. Their subsequent work, Cross-SPECL, introduces a non-adversarial ``Stabilize-then-Disentangle'' framework that exploits spectral stability through causal learning, effectively uncovering causal structure and pruning spurious features for robust cross-scene classification~\cite{chen2026crossspecl}. Data augmentation strategies designed explicitly for RS have also evolved to support generalization, including spectral variation simulation, atmospheric effect modeling, and seasonal change synthesis~\cite{zheng2024open}.}

\textcolor{black}{\textbf{Open-Set Classification.}}
\textcolor{black}{Open-set classification addresses the realistic scenario where test samples may belong to categories not seen during training, which is particularly relevant for RS applications given the vast diversity of land-cover types across different geographic regions. Traditional closed-set classifiers force all inputs into known categories, leading to systematic misclassification of novel scenes. Liu et al. proposed the ILOSR framework that combines incremental learning with open-set recognition, enabling RS models to identify unknown classes from streaming data and progressively learn new categories while mitigating catastrophic forgetting~\cite{liu2022ilosr}. For hyperspectral imagery, Ji et al. developed the Spectral-Spatial Evidential Learning Network (SSEL) that combines generative adversarial networks with evidential theory to quantify prediction uncertainty, enabling robust recognition of both known and out-of-distribution unknown samples~\cite{ji2024ssel}. These open-set methods are crucial for operational RS systems deployed in dynamic environments where novel scene types may emerge over time.}

\subsection{Attention Mechanisms in CNN Architectures}
Attention mechanisms enable CNN models to focus on discriminative regions while suppressing irrelevant information, making them particularly valuable for handling intra-class variability and inter-class similarity in RS scenes~\cite{alhichri2021classification}.

\textbf{Channel and Spatial Attention for Scene Classification.}
Channel attention mechanisms prove especially valuable for RS scene classification by adaptively recalibrating feature channels based on their importance for scene understanding. The EfficientNet-B3 with attention mechanism demonstrates significant improvements in RS scene classification by selectively emphasizing relevant feature channels while suppressing less important ones~\cite{alhichri2021classification}. 
Spatial attention mechanisms address the challenge where only portions of RS scenes may be relevant for classification, enabling models to focus on discriminative spatial regions while reducing the influence of background clutter common in RS imagery.
Zhao et al. developed a residual dense network incorporating channel-spatial attention for high-resolution remote sensing scene classification, demonstrating the effectiveness of combining both attention types for improved scene understanding~\cite{zhao2020residual}. 
This approach enables the model to simultaneously learn which feature channels are most important and which spatial locations deserve greater attention.

\textbf{Advanced Attention Integration in CNN Architectures.}
\textcolor{black}{Modern CNN architectures integrate attention mechanisms directly into convolutional layers and residual connections. Dense residual networks with embedded correlation attention learn multi-level feature correlations while maintaining efficient gradient flow, improving performance on NWPU-RESISC45 and AID~\cite{dai2024multi}. Multi-scale attention pooling adaptively weights features across spatial scales, addressing scale variance in RS scenes~\cite{bi2021multi}. Channel-spatial attention blocks simultaneously recalibrate feature channels and spatial locations within residual connections~\cite{zhao2020residual}, with attention-augmented CNNs achieving more effective feature learning than post-processing attention modules.}

\textcolor{black}{\subsection{Vision Transformers for Remote Sensing Scene Classification}}

\textcolor{black}{Vision Transformers enable modeling of global contextual relationships that CNNs struggle to capture, achieving state-of-the-art performance on standard benchmarks~\cite{bashmal2021deep,cheng2020remote}.}

\begin{figure}[!t]
    \centering
    \includegraphics[width=0.99\textwidth]{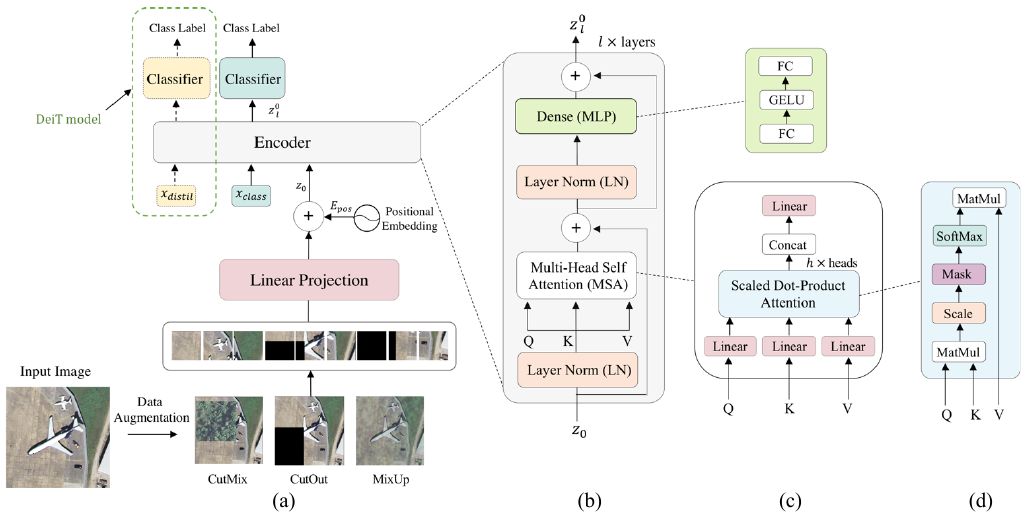}
    \caption{Vision Transformer architecture for remote sensing scene classification: (a) Overall model architecture showing patch embedding and transformer encoder stack, (b) Transformer encoder module with multi-head self-attention and MLP layers, (c) Multi-head self-attention (MSA) mechanism, (d) Individual self-attention head computation~\cite{bashmal2021deep}.}
    \label{fig:vit_architecture}
\end{figure}

\textcolor{black}{\textbf{Basic ViT Architectures and Adaptations for Scene Classification.}}
\textcolor{black}{The standard ViT architecture divides RS images into patches, embeds them with positional encodings, and applies transformer encoders to learn discriminative features for scene-level classification~\cite{bashmal2021deep}. As illustrated in Figure~\ref{fig:vit_architecture}, the ViT framework consists of patch embedding, transformer encoder blocks with multi-head self-attention, and classification heads specifically adapted for RS scene understanding. Bashmal et al. demonstrated that ViT's multi-head self-attention mechanism effectively captures contextual relations between image regions regardless of spatial distance, achieving superior performance over CNN-based methods across multiple standard datasets including UC Merced, AID, Optimal-31 and NWPU-RESISC45~\cite{bashmal2021deep}. Their compressed ViT version with pruned encoder layers maintained competitive performance, demonstrating the efficiency potential of transformer architectures for RS scene classification.}
\textcolor{black}{The Swin Transformer introduces hierarchical shifted window-based self-attention to address computational complexity while maintaining classification accuracy, proving particularly effective for multi-scale feature extraction in diverse RS scenes~\cite{liu2021swin}. LeMeViT incorporates learnable meta tokens to create sparse representations, significantly improving inference speed for RS scene interpretation while maintaining classification performance~\cite{jiang2024lemevit}.}

\textcolor{black}{\textbf{Advanced ViT Variants for Scene Classification.}}
\textcolor{black}{Multi-scale Vision Transformers address the challenge of varying object sizes in RS scenes through specialized architectures. MS-ViT employs parallel branches processing different patch sizes and fuses multi-scale features via cross-attention mechanisms, demonstrating superior performance on complex multi-label scene classification benchmarks~\cite{sun2024transformer}. Neural Architecture Search approaches have been applied to automate network design for RS scene classification, with evolutionary NAS methods outperforming manually designed architectures while using fewer parameters~\cite{chen2025evolutionary}.}
\textcolor{black}{ATMformer addresses computational complexity through adaptive token merging strategies that preserve essential local features by estimating importance scores for each token, maintaining classification performance while reducing computational overhead on standard RS scene classification benchmarks~\cite{niu2025atmformer}.}

\textcolor{black}{\subsection{State Space Models and Mamba Architectures}}
\textcolor{black}{State Space Models (SSMs), particularly the Mamba architecture, have emerged as a promising alternative to transformers for RS scene classification, offering linear computational complexity while maintaining global context modeling capabilities.}
\textcolor{black}{RSMamba introduced the first SSM-based architecture specifically designed for RS image classification, incorporating dynamic multi-path activation to handle the non-causal nature of image data~\cite{chen2024rsmamba}. By scanning images along multiple spatial directions, RSMamba captures comprehensive spatial relationships while achieving significantly lower computational overhead compared to transformer-based counterparts. This architecture demonstrates competitive performance on standard RS scene classification benchmarks including UC Merced, AID, and NWPU-RESISC45, establishing SSMs as a viable paradigm for efficient RS scene understanding.}
\textcolor{black}{The integration of Mamba with existing architectures represents an active research frontier, with hybrid Mamba-Transformer and Mamba-CNN designs showing potential for balancing local feature extraction with efficient long-range dependency modeling in complex RS scenes.}

\textcolor{black}{To summarize the deep learning methods reviewed in this section, each architecture offers distinct trade-offs for RS scene classification. CNNs excel at capturing local spatial features with well-established training procedures and moderate computational costs, making them the most widely adopted baseline. GNNs uniquely model non-Euclidean spatial relationships and inter-region dependencies, but require graph construction that adds complexity. Attention mechanisms enhance CNN representations by selectively focusing on discriminative regions, though they increase parameter counts. Vision Transformers capture global long-range dependencies that CNNs miss, achieving state-of-the-art accuracy on large-scale benchmarks, but require substantial training data and exhibit quadratic computational complexity with respect to sequence length. Mamba architectures offer a compelling alternative with linear complexity while preserving global context modeling, though they remain nascent with limited empirical validation in RS. In practice, the choice of architecture depends on the specific application constraints: data availability, computational budget, the importance of multi-scale versus global features, and whether the task requires explicit spatial relationship modeling.}

\section{\textcolor{black}{Large-Scale Pre-trained Models for Remote Sensing}}

\textcolor{black}{Large-scale pre-trained foundation models leverage massive unlabeled datasets through self-supervised learning, achieving remarkable generalization and zero-shot classification capabilities~\cite{sun2022ringmo,guo2024skysense}. Characterized by billion-scale parameters and pre-training on vast RS imagery, they differ fundamentally from the deep learning architectures in Section~4. This section covers two major paradigms: Remote Sensing Foundation Models (RSFMs) and Vision-Language Models (VLMs).}

\subsection{Remote Sensing Foundation Models for Scene Classification}

Remote Sensing Foundation Models (RSFMs) are large-scale pretrained models designed for downstream RS tasks including scene classification~\cite{sun2022ringmo,guo2024skysense}. They leverage vast unlabeled satellite imagery to address annotation scarcity and improve zero-shot classification across diverse geographic regions and sensor types.

\textbf{Large-Scale Foundation Model Architectures.} 
SkySense stands as the largest multi-modal RSFM to date, featuring billion-scale parameters pretrained on extensive temporal sequences of optical and SAR imagery~\cite{guo2024skysense}. The model incorporates a factorized multi-modal spatiotemporal encoder trained through Multi-Granularity Contrastive Learning, enabling robust scene classification across different sensor modalities. SkySense demonstrates remarkable generalization on scene classification benchmarks, achieving strong performance through fine-tuning only classification heads while freezing encoder layers~\cite{guo2024skysense}.
RingMo employs masked image modeling with ring-based masking strategies to learn rotation-invariant features particularly valuable for RS scene classification, where scenes may appear at arbitrary orientations~\cite{sun2022ringmo}. The model excels in cross-domain transfer scenarios, demonstrating strong performance when transferring between different satellite sensors and geographic regions.

\textbf{Self-Supervised Learning Strategies for Scene Classification.} 
\textcolor{black}{Self-supervised learning uses pretext tasks to learn meaningful representations without manual annotations. SSL4EO applies contrastive learning on large-scale Sentinel imagery, demonstrating strong linear probing performance for seasonal scene variations~\cite{wang2023ssl4eo}. Scale-MAE handles multi-scale geospatial representations by reconstructing masked image patches at different resolutions, excelling in few-shot scene classification~\cite{reed2023scale}. Parameter-efficient fine-tuning adapts RSFMs for scene classification datasets with competitive performance at reduced computational cost~\cite{guo2024skysense}.}

\subsection{Vision-Language Models for Scene Classification}

Vision-Language Models (VLMs) integrate visual and textual modalities to enable semantic-rich RS scene classification, supporting natural language queries and zero-shot classification of previously unseen scene categories~\cite{liu2024remoteclip,li2023rs}.

\textbf{CLIP-Based Models for Remote Sensing Scene Classification.} 
RemoteCLIP represents the first comprehensive vision-language foundation model specifically designed for remote sensing applications, including scene classification~\cite{liu2024remoteclip}. Through data scaling techniques that convert heterogeneous annotations into unified image-caption formats, RemoteCLIP creates a significantly larger pretraining dataset than existing approaches. The model demonstrates superior zero-shot scene classification performance, substantially outperforming CLIP baselines across multiple downstream scene classification datasets~\cite{liu2024remoteclip}.
RS-CLIP introduces pseudo-labeling techniques for zero-shot scene classification, achieving remarkable performance on novel scene categories across standard benchmarks including UC Merced, WHU-RS19, NWPU-RESISC45, and AID datasets~\cite{li2023rs}. The approach incorporates curriculum learning strategies that progressively improve model performance through multiple stages of fine-tuning.
GeoRSCLIP leverages large-scale image-text paired datasets to bridge the gap between general vision-language models and domain-specific RS scene classification tasks~\cite{zhang2024rs5m}. The model incorporates geographic context through location-aware prompts, improving semantic understanding of region-specific scene characteristics.

\textbf{Text-Guided Scene Classification and Interpretation.} 
Vision-language models enable semantic reasoning over RS scenes through natural language descriptions, facilitating interpretable scene classification results. These models support complex queries like ``urban areas with vegetation" or ``agricultural regions near water bodies," enabling more nuanced scene understanding beyond traditional fixed-category classification~\cite{zhang2024rs5m}.
The integration of vision-language models with generative capabilities enables explainable scene classification outputs, where models can provide natural language justifications for their classification decisions, crucial for applications in environmental monitoring and urban planning~\cite{huang2025textscd}.
Contemporary VLM developments demonstrate strong performance in cross-modal scene retrieval tasks, where scenes can be classified or retrieved based on textual descriptions, opening new possibilities for large-scale geographic information systems and automated Earth observation analysis~\cite{liu2024remoteclip}.

\textcolor{black}{Comparing the two paradigms reviewed in this section, RSFMs and VLMs address different aspects of the data-scarcity challenge. RSFMs achieve strong performance through self-supervised visual pre-training on large-scale unlabeled RS imagery, excelling in fine-tuning and few-shot scenarios where task-specific labels are limited but visual representations can be effectively transferred. VLMs, by contrast, leverage vision-language alignment to enable zero-shot classification without any task-specific training, and provide interpretable outputs through natural language. However, RSFMs generally achieve higher accuracy on standard benchmarks when fine-tuning data is available, while VLMs are more flexible for open-vocabulary tasks and novel category discovery. Both paradigms incur substantial computational costs during pre-training, and their effective deployment to operational RS systems remains an active area of research.}

\section{Generative AI for Scene Classification Enhancement}

Generative AI addresses data scarcity, class imbalance, and domain adaptation challenges in RS scene classification~\cite{cheng2017remote}. Note that direct application of generative models to scene-level classification remains nascent—most existing work targets pixel-level tasks such as semantic segmentation or image restoration~\cite{dash2023review}. This section focuses on contributions directly relevant to scene-level understanding.

\subsection{Generative Adversarial Networks Applications}

\textbf{GAN-Based Data Augmentation for Scene Classification.}
\textcolor{black}{Direct GAN applications for scene classification augmentation remain sparse. Modified shuffle attention GAN approaches show promise for generating higher-quality images with limited inputs~\cite{chen2021remote}, but comprehensive evaluation on standard scene classification benchmarks (UCM, AID, NWPU-RESISC45) remains limited. Semi-supervised GAN frameworks incorporate unlabeled data to reduce annotation costs~\cite{kwak2023semi}, with GANs identified as promising solutions for class imbalance~\cite{hao2023review}, though systematic evaluation is still needed.}

\textbf{Cross-Domain Adaptation via Adversarial Training.}
Adversarial training approaches for domain adaptation address the cross-domain challenges inherent in RS scene classification when deploying models across different geographic regions or sensor types~\cite{voreiter2020cycle,soto2020domain}. The cycle-consistent adversarial training framework enables unpaired image-to-image translation between domains where paired data is unavailable, providing a foundation for cross-sensor and cross-region scene classification adaptation.

\subsection{Variational Autoencoders and Diffusion Models}

\textbf{VAEs for Scene-Level Representation Learning.}
VAEs have been incorporated into remote sensing image understanding systems, with applications including scene captioning that demonstrate the utility of VAE-learned latent representations for scene content understanding~\cite{shen2020remote}. This represents a direct connection between VAE-based generative modeling and scene-level semantic understanding in remote sensing, where the learned latent space captures holistic scene characteristics useful for classification.

\textbf{Diffusion Models for Scene Understanding.}
Diffusion models show emerging potential for RS scene classification, particularly through their ability to learn rich feature representations from unlabeled data. A notable direct application is DiffPR-Net, published in IEEE Transactions on Geoscience and Remote Sensing in 2025, which addresses few-shot remote sensing scene classification by incorporating a diffusion augmentation module that generates high-quality synthetic scene images to expand limited support sets~\cite{zhu2025diffprnet}. The model combines this generative augmentation with a dual attention fusion module and a prototype rectified module, achieving competitive performance on standard few-shot scene classification benchmarks and demonstrating that diffusion-based data synthesis can directly benefit scene-level recognition under data-scarce conditions.
Multimodal classification approaches using Masked Diffusion Spatio-Temporal Autoencoders (MDSTA) address missing modality challenges through conditional masked diffusion processes, demonstrating relevance to scene-level understanding with multimodal remote sensing data~\cite{yue2025mdsta}.
Diffusion model applications in remote sensing have been systematically reviewed, revealing extensive use for image generation, enhancement, and interpretation tasks, with scene classification applications beginning to emerge~\cite{liu2024diffusion}. The integration of diffusion models with transformer architectures for unsupervised hierarchical feature learning provides promising foundations for scene classification~\cite{sigger2024unveiling,luo2024rs}.

Current challenges for generative AI in scene classification include the complexity of RS scene patterns, computational requirements for operational deployment, and the absence of systematic evaluation protocols for synthetic data effectiveness on standard scene classification benchmarks. Key future directions include developing scene-specific generative models, investigating generative approaches for few-shot scene classification, and exploring integration with foundation models such as RingMo~\cite{sun2022ringmo} and SkySense~\cite{guo2024skysense} for enhanced scene understanding through synthetic data pretraining.

\textcolor{black}{Among the three generative paradigms, GANs produce the most visually realistic synthetic RS images through adversarial training but suffer from training instability and mode collapse, potentially limiting the diversity of generated samples. VAEs offer more stable training and principled latent space modeling, yet tend to produce blurrier outputs that may be less effective for augmenting high-resolution RS datasets. Diffusion models achieve the highest generation quality and diversity, and their iterative denoising process naturally captures multi-scale features, but they incur significantly higher computational costs during both training and inference. For practical RS scene classification augmentation, GANs remain the most widely adopted due to their balance of generation quality and efficiency, while diffusion models represent the most promising future direction as computational constraints are addressed.}

\section{Hybrid Approaches and Real-World Applications}

Hybrid approaches combine multiple architectural paradigms and specialized methodologies to address contemporary RS scene classification challenges.

\subsection{CNN-Transformer Hybrid Architectures}

\textbf{Plug-and-Play Hybrid Networks.} P2FEViT embeds CNN features directly into Vision Transformer architectures, integrating local feature extraction with global context modeling while mitigating ViT's data requirements~\cite{wang2023p}. The approach demonstrates effectiveness in both complex scene classification and fine-grained recognition tasks with improved convergence on limited training data.
Aboghanem et al. proposed hybrid models combining ResNet-50 and Vision Transformer features with multi-head attention mechanisms for aerial image classification, demonstrating that CNN-Transformer fusion with attention-based feature selection achieves competitive performance on standard RS benchmarks~\cite{aboghanem2026hybrid}.

\textbf{Lightweight Ensemble Frameworks.} The Exceptionally Straightforward Ensemble (ESE) method achieves enhanced performance through pure data correction rather than complex architectural modifications~\cite{song2025pure}. This approach implements quantitative augmentation via plug-and-play modules that effectively correct feature distributions across remote sensing data. The method demonstrates superior performance compared to numerous recent techniques while maintaining computational efficiency through lightweight ensemble strategies that combine CNN and Vision Transformer models.

\textbf{Multi-Scale Feature Integration.} Multi-scale dense residual correlation networks integrate multi-stream feature extraction modules with dense residual connected feature fusion technology and Correlation Attention Modules~\cite{dai2024multi}. These architectures learn feature representations at multiple levels while addressing the complex spatial organization characteristic of remote sensing scenes, demonstrating effectiveness across standard benchmarks.

\subsection{Advanced Transformer Variants and Specialized Learning Paradigms}

\textbf{Adaptive Token Management.} Adaptive token management reduces computational overhead by selectively merging or pruning less informative tokens while preserving discriminative ones, achieving significant speedups without sacrificing accuracy and enabling practical deployment on standard hardware~\cite{niu2025atmformer}.

\textbf{Few-Shot Learning Integration.} Few-shot learning approaches enable effective scene classification with limited labeled samples through meta-learning frameworks. RS-MetaNet employs metric learning for prototype-based adaptation, achieving competitive performance in scenarios with scarce training data~\cite{li2020rs}. Heterogeneous prototype distillation with support-query correlative guidance further advances few-shot RS scene classification by distilling discriminative prototypes across heterogeneous feature spaces~\cite{li2024heterogeneous}. Visual prompt tuning approaches on Vision Transformers enhance zero-shot and few-shot performance through efficient adaptation of pre-trained foundation models~\cite{zhu2024mvp}.

\textbf{Self-Supervised Contrastive Learning.} Self-supervised learning methods pre-train models on unlabeled remote sensing data using contrastive objectives. Multiview contrastive coding approaches learn invariant representations from different augmentations of the same scene~\cite{stojnic2021self}. Spatial-temporal contrastive learning handles multi-temporal data by learning consistent representations across time, addressing temporal dependencies in dynamic scene understanding~\cite{huang2022spatial}.

\textbf{Active Learning Systems.} Active learning frameworks iteratively select informative samples for annotation, incorporating human expertise to reduce labeling costs. Graph-based active learning approaches achieve significant annotation cost reductions while maintaining classification performance~\cite{miller2022graph}. Interactive learning systems query uncertain scenes and incorporate human feedback for improved model adaptation~\cite{lenczner2022dial}.

\textbf{Continual and Incremental Learning.} Continual learning addresses the challenge of adapting scene classification models to new classes or domains over time without catastrophic forgetting of previously learned knowledge. The CLRS benchmark established a standardized evaluation framework for continual learning in RS scene classification, enabling systematic comparison of rehearsal, regularization, and architecture-based strategies~\cite{li2020clrs}. Dual knowledge distillation with classifier discrepancy has been proposed for incremental scene classification, balancing stability and plasticity when learning new scene categories on both natural and remote sensing images~\cite{yu2024incremental}. Domain incremental learning, where models must adapt to data from new geographic regions or sensor configurations, has been addressed through contrastive dual-pool feature adaptation that maintains discriminative representations across evolving domains~\cite{shao2025contrastive}.

\textbf{Semi-Supervised and Noisy-Label Learning.} Semi-supervised learning methods leverage large amounts of unlabeled RS imagery alongside limited labeled samples to improve scene classification performance. These approaches are particularly valuable given the high annotation costs associated with RS scene data, where expert knowledge is often required for accurate labeling~\cite{guo2024scene}. Noisy label distillation addresses the practical challenge of label noise in RS scene datasets, where annotation errors arise from ambiguous scene boundaries and subjective categorization, by training robust models that learn to identify and mitigate the effects of incorrect labels~\cite{zhang2020noisy}.

\subsection{Real-World Applications and Deployment Challenges}

\textbf{Environmental Monitoring and Urban Planning.} Remote sensing scene classification supports urban planning through automated analysis of high-resolution imagery for land-use change monitoring. Applications include classifying functional urban zones such as residential, commercial, and industrial areas for sustainable development planning~\cite{xie2022land}. Environmental monitoring leverages multimodal approaches for ecosystem health assessment and climate analysis through comprehensive land cover classification~\cite{cai2024multimodal}.

\textbf{Agricultural Applications.} Cross-domain few-shot learning enables crop type mapping across different geographic regions with minimal training data, facilitating scalable agricultural monitoring systems~\cite{li2025agrifm}. Scene classification approaches support precision agriculture through automated analysis of agricultural landscapes and crop health assessment across diverse growing conditions.

\textbf{Disaster Management and Emergency Response.} Scene classification enables rapid damage assessment following natural disasters through automated analysis of post-event imagery. Classification systems support disaster response coordination by identifying affected areas and infrastructure damage across large geographic regions. Real-time processing capabilities are essential for timely emergency response and resource allocation decisions.

\textbf{Edge Computing and Federated Learning Deployment.} Recent developments focus on creating classification frameworks suitable for edge deployment, emphasizing parameter efficiency and computational optimization for real-time applications~\cite{sitaula2024enhanced}. Federated learning approaches enable collaborative model training across distributed data sources while preserving data privacy, particularly valuable for applications involving sensitive geographic information~\cite{ben2024federated}.

\textbf{Cross-Domain Generalization Challenges.} Operational deployment requires robust performance across different geographic regions, sensor configurations, and environmental conditions. Advanced domain adaptation methods address these challenges by learning domain-invariant representations while maintaining discriminative capability for scene classification~\cite{wang2022transferring}. These approaches are crucial for developing scalable remote sensing systems that can operate effectively across diverse deployment scenarios without requiring extensive retraining.

\textbf{Scalability and Computational Efficiency.} Real-world applications demand careful balance between classification accuracy and computational requirements. Lightweight architectures and model compression techniques enable deployment on resource-constrained platforms while maintaining acceptable performance levels. Edge computing solutions facilitate on-device processing for applications requiring immediate response without network connectivity dependencies.

\section{Challenges, Emerging Trends, and Future Directions}

While remote sensing scene classification has achieved remarkable progress with deep learning and foundation models, the field faces new challenges and opportunities shaped by evolving technological capabilities, deployment requirements, and societal needs. This section addresses contemporary challenges beyond the fundamental technical issues discussed in Section~\ref{sec:frssc}, explores emerging technological trends, and identifies future research directions.

\subsection{Contemporary Challenges}

Contemporary challenges include deployment constraints, model interpretability, and ethical considerations that extend beyond the fundamental technical issues of Section~\ref{sec:frssc}.

\textbf{Model Interpretability and Explainability.}
Interpretability remains critical for RS scene classification applications in disaster management, environmental monitoring, and policy-making. Most explainable AI methods and related evaluation metrics in RS are initially developed for natural images considered in computer vision, and their direct usage in RS may not be suitable~\cite{klotz2025effectiveness}.
A systematic review on explainable AI in remote sensing reveals that deep neural networks, despite achieving state-of-the-art performance for RS tasks such as scene classification, are often considered black-box solutions with prediction processes that are not well understood~\cite{hohl2024opening}.
The lack of understanding may lead to biased predictions or limitations of model performances, potentially resulting in erroneous decision-making~\cite{hohl2024opening}.
Current XAI methods face specific evaluation challenges in RS scene classification, where classification often relies more on spectral and texture features distributed across multiple regions, making traditional explanation methods less suitable~\cite{klotz2025effectiveness}.
\textcolor{black}{Several XAI techniques have been applied to RS scene classification with varying degrees of success. Grad-CAM generates class-discriminative saliency maps by computing gradients of the predicted class score with respect to convolutional feature maps, revealing which spatial regions drive CNN-based classification decisions. LIME (Local Interpretable Model-agnostic Explanations) explains individual predictions by perturbing superpixel-level inputs and fitting local interpretable models, offering insight into which scene components contribute to specific class assignments. SHAP (SHapley Additive exPlanations) provides theoretically grounded feature attribution scores based on cooperative game theory, quantifying the contribution of each input feature to model predictions. However, the standard evaluation metrics used to assess these XAI methods in the natural image domain often fail when applied to RS scene classification, due to fundamental differences in image characteristics. \textit{Insertion/Deletion AUC} metrics evaluate explanations by progressively adding or removing pixels ranked by importance and measuring classification confidence changes; these metrics implicitly assume that individual pixels carry independent semantic meaning---an assumption severely violated in RS imagery, where meaningful features are defined by spatial arrangements, texture patterns, and multi-scale structures rather than individual pixel values. \textit{Pointing Game accuracy} measures whether the most salient point falls within the ground-truth object region, assuming a single localized discriminative region; this assumption is inappropriate for RS scenes that contain multiple semantically relevant objects at vastly different scales (e.g., individual vehicles alongside entire runways in an airport scene, or scattered buildings defining a residential area). \textit{IoU-based evaluation} metrics require ground-truth saliency annotations, which are rarely available for RS scene classification because scene-level labels do not inherently specify which spatial regions are discriminative. The root causes of these failures are intrinsically linked to the spatial and physical characteristics of RS imagery: (i)~\textit{multi-scale objects}---RS scenes contain objects spanning orders of magnitude in size, from individual trees to entire forest blocks, making single-scale saliency maps inherently incomplete; (ii)~\textit{highly complex backgrounds}---unlike natural images where backgrounds are typically homogeneous, RS backgrounds exhibit rich spatial texture that may itself be diagnostic for scene classification, causing attribution methods to misidentify salient regions; (iii)~\textit{spectral feature dependence}---RS scene classification often relies on spectral signatures across multiple bands that are not captured by RGB-based saliency visualization; and (iv)~\textit{distributed discriminative information}---the features defining an RS scene category are typically distributed across the entire image rather than concentrated in a single region, fundamentally challenging localization-based explanation paradigms.}

\textbf{Scalability and Real-World Deployment Constraints.}
\textcolor{black}{RS image classification is challenging due to data complexity, diversity, and sparsity~\cite{song2025pure}. Manual annotation demands substantial expertise and incurs high costs, limiting dataset scalability~\cite{cheng2020remote,guo2024scene}. Effective integration of multimodal Earth observation data remains underexplored~\cite{xiao2025foundation}, and existing methods often require substantial architectural modifications~\cite{song2025pure}.}

\textbf{Ethical Considerations and Model Bias.}
Systematic biases in RS scene classification arise from training data characteristics and model assumptions. Issues such as limited labeled samples and class imbalance may lead to classification bias in classifiers, with data scarcity and class imbalance representing ongoing challenges that need to be addressed by remote sensing communities~\cite{tombe2023remote}. 
The development of accurate methods for RS scene classification has been dominated by deep learning approaches that extract and exploit complex spatial and spectral content. Yet, these models can contain multiple land-use land-cover classes simultaneously, creating multi-label classification challenges that traditional single-label approaches cannot adequately address~\cite{klotz2025effectiveness}.
The need to address data privacy, algorithmic fairness, and the environmental impact of large-scale model training has become increasingly crucial for operational RS systems~\cite{paolanti2024ethical}.

\textbf{Adversarial Robustness.}
Deep learning models for RS scene classification are vulnerable to adversarial perturbations that can cause incorrect predictions with high confidence, posing significant risks in safety-critical applications such as disaster response and military reconnaissance. A comprehensive study on the robustness of deep learning-based image classification in remote sensing reveals that DNNs are susceptible to various adversarial attacks including universal perturbations, patch-based attacks, and physically realizable manipulations~\cite{xu2024adversarial}. Developing robust scene classification models that maintain reliable performance under adversarial conditions remains an open challenge, requiring advances in adversarial training, certified defenses, and robustness evaluation frameworks tailored to the unique characteristics of RS imagery.

\subsection{Emerging Trends}
Emerging trends in RS scene classification leverage computational advancements, novel architectures, and interdisciplinary approaches to enable real-time, scalable, and sustainable solutions.

\subsubsection{Edge Computing, Real-Time Processing, and Federated Learning}
\textbf{Edge Computing and Real-Time Processing.}
\textcolor{black}{Edge computing enables on-device RS scene classification, crucial for applications requiring immediate response, by addressing latency and bandwidth constraints~\cite{moreno2024federated}. Challenges include inconsistent image resolution, environmental interference, scene class imbalance, and limited compute resources, requiring lightweight solutions for edge deployment~\cite{sitaula2024enhanced}.}

\textbf{Federated Learning.} 
\textcolor{black}{Federated learning enables collaborative model training across distributed sources without data centralization~\cite{ben2024federated}. FedRSCLIP, the first federated VLM framework for RS, addresses data heterogeneity and model transmission challenges~\cite{lin2025fedrsclip}. Advanced federated approaches further introduce explanation-guided pruning to reduce model complexity while improving communication efficiency~\cite{klotz2025communication}.}

\subsubsection{Vision-Language Models, Cross-Modal Fusion, and Sustainable AI}
\textbf{Vision-Language Models.} 
\textcolor{black}{Vision-Language Models enable semantic reasoning and natural language interaction for RS scene classification. FedRSCLIP uses dual-prompt mechanisms—shared prompts for global knowledge and private prompts for client-specific adaptation—bridging federated learning with VLM pretraining~\cite{lin2025fedrsclip}. Recent advances establish new baselines for few-shot scene classification by harnessing large VLMs as image encoders~\cite{dutta2023remote}.}

\textbf{Cross-Modal Fusion.} 
\textcolor{black}{Cross-modal fusion integrates diverse sensor modalities for robust scene understanding. Recent work demonstrates effective fusion of hyperspectral and RGB imagery, surpassing traditional methods in accuracy and robustness~\cite{li2024stemnet}. Multi-view multi-source transfer learning leverages diverse EO data to improve generalization~\cite{paheding2024advancing}.}

\textbf{Sustainable AI.} 
\textcolor{black}{Sustainable AI reduces computational and energy requirements through lightweight ensemble methods, model pruning, and parameter-efficient fine-tuning, enabling deployment on resource-constrained edge devices~\cite{song2025pure,klotz2025communication}.}

\subsection{Future Research Directions}
Future research directions should address current limitations while capitalizing on emerging technological capabilities.

\subsubsection{Scaling for Advanced Data Modalities and Multi-Temporal Analysis}

\textbf{Advanced Data Modalities.} 
Future research should focus on developing domain-specific architectures optimized for diverse remote sensing data types, leveraging multi-source data from multiple perspectives to improve cross-domain generalization~\cite{paheding2024advancing}. 
Research opportunities include developing specialized models for hyperspectral-spatial correlations and self-supervised learning approaches for multi-spectral scene classification.

\textbf{Multi-Temporal Analysis.} 
Multi-temporal analysis requires models capturing complex temporal dynamics across diverse landscapes. Future research should develop memory-efficient temporal models and comprehensive benchmark datasets for multi-temporal scene analysis.

\subsubsection{Hardware Innovation, Model Optimization, and Standardized Benchmarking}

\textbf{Hardware Innovation and Model Optimization.} 
Specialized hardware development focuses on enabling real-time scene classification processing on resource-constrained devices. Model optimization techniques show significant promise for RS scene classification, with lightweight architectures addressing complex spatial organization challenges while balancing computational efficiency with feature extraction capability~\cite{liu2025efficient}. 
Privacy-preserving and efficient transfer learning, such as developing transfer learning models in federated learning settings that are robust to adversarial attacks and compatible with varying computational resources, represent critical future directions.
Neural architecture search approaches designed explicitly for RS scene classification tasks present opportunities for automated model design.

\textbf{Standardized Benchmarking.} 
Standardized benchmarking remains crucial for fair model comparison in RS scene classification. Current evaluation approaches vary significantly across different datasets and methods, limiting reproducibility and progress assessment~\cite{tombe2023remote}. 
Comprehensive benchmark suites addressing traditional scene classification and emerging challenges like few-shot learning, cross-domain adaptation, and federated learning scenarios could accelerate research progress.

\textbf{Towards Trustworthy and Responsible AI.}
Future developments should prioritize explainability, fairness, and environmental sustainability. Explainable AI is crucial for high-stakes applications such as disaster response and environmental monitoring~\cite{janga2023review}, requiring interpretable foundation models, ethical guidelines for automated Earth observation, and evaluation frameworks assessing both technical and societal impact~\cite{paolanti2024ethical}.

\textbf{Uncertainty Quantification and Robustness.}
Reliable uncertainty estimation in scene classification predictions is essential for operational deployment, particularly in high-stakes applications where overconfident misclassifications can lead to costly errors. Future research should develop calibrated uncertainty quantification methods that distinguish between aleatoric uncertainty arising from inherent scene ambiguity and epistemic uncertainty due to limited training data or out-of-distribution inputs. Integrating uncertainty-aware frameworks with adversarial robustness techniques~\cite{xu2024adversarial} and continual learning paradigms~\cite{shao2025contrastive} represents a promising direction toward building trustworthy RS scene classification systems capable of flagging unreliable predictions for human review.

\section{Conclusion}
\textcolor{black}{This survey traces RS scene classification evolution from traditional handcrafted methods to cutting-edge generative AI and foundation models.}

\textbf{Summary of Key Findings and Evolution of Methods.}
\textcolor{black}{The field evolved from manual feature engineering to automated learning. Traditional handcrafted features (GLCM, SIFT) with classical classifiers faced scalability limitations; deep CNNs enabled end-to-end learning with significant accuracy gains; ViTs and foundation models (RingMo, SkySense) achieved state-of-the-art performance through global modeling and self-supervised pretraining; generative AI and hybrid architectures address data scarcity and computational efficiency.}

\textbf{Implications for Researchers and Practitioners.}
\textcolor{black}{Researchers should prioritize multimodal fusion, interpretability, and lightweight architectures; self-supervised learning enables few-shot and cross-domain adaptation. Practitioners benefit from edge computing solutions and VLMs for rapid decision-making, while foundation models reduce annotation requirements across diverse deployments.}

\textbf{Final Remarks and Call for Interdisciplinary Collaboration.}
RS scene classification has matured into an AI-driven field offering unprecedented Earth observation capabilities. Despite progress, challenges in data scarcity and computational efficiency require continued innovation. The convergence of generative AI, foundation models, and edge computing creates opportunities for scalable Earth monitoring systems. We call for interdisciplinary collaboration to develop ethical, efficient AI solutions through open-source models and standardized frameworks, ensuring equitable access to advanced RS technologies for addressing global challenges.

\backmatter














\bibliography{refs.bib}

\end{document}